\documentclass[conference]{IEEEtran}
\IEEEoverridecommandlockouts
\usepackage{cite}
\usepackage{amsmath,amssymb,amsfonts}
\usepackage{algorithmic}
\usepackage{graphicx}
\usepackage{textcomp}
\usepackage{xcolor}
\def\BibTeX{{\rm B\kern-.05em{\sc i\kern-.025em b}\kern-.08em
    T\kern-.1667em\lower.7ex\hbox{E}\kern-.125emX}}

\begin{document}

\title{Real-time Mortality Prediction Using MIMIC-IV ICU Data Via Boosted Nonparametric Hazards

\thanks{This work was supported in part by NIH grant 1R21EB028486-01. \\ Published in: Nowroozilarki Z, Pakbin A, Royalty J, Lee DK, Mortazavi BJ. Real-time Mortality Prediction Using MIMIC-IV ICU Data Via Boosted Nonparametric Hazards. In 2021 IEEE EMBS International Conference on Biomedical and Health Informatics (BHI) 2021 Jul 27 (pp. 1-4). IEEE.}
}

\author{\IEEEauthorblockN{Zhale Nowroozilarki\IEEEauthorrefmark{1}, Arash Pakbin\IEEEauthorrefmark{1}, James Royalty\IEEEauthorrefmark{1}, Donald K.K. Lee\IEEEauthorrefmark{2}, and Bobak J. Mortazavi\IEEEauthorrefmark{1}} \IEEEauthorblockA{\IEEEauthorrefmark{1}Department of Computer Science \& Engineering, Texas A\&M University,\\Email:  \{zhale, a.pakbin, troyalty18, bobakm\}@tamu.edu} \IEEEauthorblockA{\IEEEauthorrefmark{2}Goizueta Business School and Dept of Biostatistics \& Bioinformatics, Emory University, Email: donald.lee@emory.edu}
}

\maketitle

\begin{abstract}
Electronic Health Record (EHR) systems provide critical, rich and valuable information at high frequency. One of the most exciting applications of EHR data is in developing a real-time mortality warning system with tools from survival analysis. However, most of the survival analysis methods used recently are based on (semi)parametric models using static covariates. These models do not take advantage of the information conveyed by the time-varying EHR data.
In this work we present an application of a highly scalable survival analysis method, BoXHED 2.0 \cite{pakbin2021boxhed}, to develop a real-time in-ICU mortality warning indicator based on the MIMIC IV data set \cite{johnson2016mimic}. Importantly, BoXHED can incorporate time-dependent covariates in a  fully nonparametric manner and is backed by theory \cite{LCI}. Our in-ICU mortality model achieves an AUC-PRC of 0.41 and AUC-ROC of 0.83 out of sample, demonstrating the benefit of real-time monitoring. 
\end{abstract}


\section{Introduction}
Electronic Health Records (EHRs) and other health information technology (e.g. personal data from wearable sensing) provide a potential trove of data for clinical risk modeling. Its real-time nature represent a major advantage over administrative or registry-based data \cite{schulz2020agile}. However, the development of clinical models tends to remain within large registries that abstract the data into static snapshots of patient health \cite{mcnamara2016predicting}. Even machine learning models fail to substantively improve the prediction on these time-static datasets \cite{khera2021use}, highlighting the need for methods that leverage richness of the EHR data to improve performance \cite{engelhard2021incremental}.

With the availability of ICU data in the form of the MIMIC-III and MIMIC-IV datasets \cite{johnson2020mimic}, models that take advantage of EHR data have gained prominence. While these models show promise in predicting important clinical outcomes such as mortality \cite{harutyunyan2019multitask, pakbin2018prediction}, they tend to only generate one classification prediction at one point in time during the entire episode of care. For example, \cite{mortazavi2017prediction} uses the first 24 hours of data to predict outcomes after cardiovascular procedures. Prediction systems for in-ICU mortality that are more dynamic in nature update forecasts periodically using the latest information available \cite{ma2019shapes,wang2020mimic}.

Ideally, adverse event warning systems should operate in real-time as a patient's episode of care evolves. An example of this can be seen in the prediction of sepsis in admitted patients \cite{henry2015targeted}. However, existing real-time prediction methodologies are based on classical statistical models that do not take advantage of recent advances in machine learning. The purpose of this paper is to explore the performance improvement that can be gained from embedding state-of-the-art survival analysis techniques into real-time mortality warning systems.

We focus on a recent survival methodology called BoXHED \cite{wang2020boxhed, pakbin2021boxhed}, which is a gradient boosted procedure that is well suited to estimating clinical risk in the presence of time-varying features from EHRs. We train BoXHED to the MIMIC-IV dataset, and use it to create an in-ICU mortality warning system that continually assesses risk as the features evolve. This is particularly relevant to patients with short stays (under 5 days) as they have the most variable conditions. Out-of-sample performances are compared to two benchmarks: The classic time-varying Cox model, and a method that is representative of recent deep learning approaches for survival data (Dynamic DeepHit).

\section{Survival models for time-varying features}
Techniques used to forecast the time $T$ to an event (e.g. mortality) fall under the survival analysis discipline. When time-dependent features $X(t)$ are involved, it is shown in \cite{wang2020boxhed} that the fundamental quantity of interest is the hazard function $\lambda(t,x)$, which is the conditional probability of the event occurring in the next instant given that it has not yet occurred:
\begin{equation}\label{eq:haz}
\lambda(t,x)dt \approx \mathbb{P}(T\in[t,t+dt) | T\ge t, X(t)=x).
\end{equation}
Thus $\lambda(t,X(t))$ is the most natural measure of real-time mortality risk. Note that $X(t)$ can either be the current values of the features, or it can be feature-engineered to be its history up to $t$.

\subsection{Cox proportional hazards model} The venerated Cox model \cite{cox1972regression} is the workhorse model used in applications, but it imposes a key assumption on the functional form of $\lambda(t,x)$ that rules out potential interaction effects between time and the covariates:
\begin{equation}\label{eq:cox}
\lambda_{PH}(t,x) = h_0(t) e^{R(x)}.
\end{equation}
The baseline hazard function $h_0(t)$ is difficult to estimate from data without further assumptions, but $R(x)$ can be estimated independently of $h_0(t)$. Thus $R(x)$ provides a relative risk score that can be used to compare subjects, but an absolute risk score is not readily available.

Traditionally, $R(x) = \beta'x$ is modeled linearly and this specification permits the inclusion of time-varying features \cite{zhang2018time}. Following the sepsis prediction application \cite{henry2015targeted}, we use $R(X(t))$ as a real-time measure of (relative) mortality risk for the Cox model benchmark.

We note that recent works in machine learning relax the linear specification by using deep neural networks to model $R(x)$ \cite{katzman2018deepsurv, nagpal2021deep}. However, these methods only accommodate time-static covariates and are therefore unable to take advantage of the information conveyed by the time-varying clinical data.

\subsection{Deep learning survival models} In a step towards accommodating time-dependent covariates in a nonparametric way, deep survival models forgo real-time prediction by specializing to discrete time $\tau$ \cite{ren2019,lee2019dynamic}. Survival prediction then becomes solving binary classification problems at each point on a predefined time grid. For example, Dynamic DeepHit \cite{lee2019dynamic} further imposes a predefined time $\tau_{\max}$ by which the event has to happen with probability one, so that a recurrent neural network with a soft-max output layer can be used to estimate
\begin{equation}\label{eq:DD}
    o_\tau := \mathbb{P}(T = \tau| \mathcal{X})
\end{equation}
for $\tau\le\tau_{\max}$ and $\sum_{\tau \le \tau_{\max}} o_\tau = 1$. Here, $\mathcal{X}$ is the covariate history up to the time of prediction. Extension to competing risks is also studied in \cite{lee2019dynamic}.

To use Dynamic DeepHit to produce a mortality risk measure at time $t$ for our deep learning benchmark, we discretize continuous time into hourly bins $\tau_1,\cdots,\tau_{\max}$ and denote the bin containing $t$ as $\tau(t)$. The risk measure at $t$ is then taken to be $o_{\tau(t)}$, which is the discrete approximation to the hazard \eqref{eq:haz} at $t$.

We note that the chief purpose of \cite{lee2019dynamic} is not to estimate \eqref{eq:DD}, but to use them to estimate the cumulative probabilities $\mathbb{P}(T\le \tau|\mathcal{X})$. For our application, the cumulative probability is not an appropriate risk measure because our prediction target is whether or not a patient dies in the ICU at \textit{any} point after $t$. Since Dynamic DeepHit assumes that the event must occur by $\tau_{\max}$ hours in the ICU ($\sum_{\tau \le \tau_{\max}} o_{\tau} = 1$), the predicted cumulative probability of in-ICU death will always be 1.

\subsection{Boosted nonparametric hazards} Very recently, \cite{LCI} developed a theoretically justified gradient boosting solution for estimating the hazard \eqref{eq:haz} nonparametrically with (continuous) time-varying features. A scalable tree-boosted implementation called BoXHED can handle recurrent events as well as survival data beyond right-censoring \cite{pakbin2021boxhed}. Support for missing data and multicore CPU/GPU computing are also included. BoXHED performs regularized minimization of the negative nonparametric likelihood, and does so by iteratively adding shallow regression trees. In contrast to the Cox model, the BoXHED hazard estimator $\hat\lambda(t,x)$ provides an absolute measure of a subject's real-time mortality risk rather than a relative one.

\section{Methods}
We compare the performance of BoXHED to those of the baselines (time-varying Cox and Dynamic DeepHit) at predicting in-ICU mortality on a continuous basis. The data comes from MIMIC IV \cite{johnson2020mimic}. We follow the approach in the sepsis prediction application \cite{henry2015targeted} to convert survival risk measures into real-time mortality predictions, which is to use the classic sliding window to update risks. While the Cox relative risk $R(X(t))$ was used in \cite{henry2015targeted}, this work evaluates the improvements brought specifically by BoXHED's boosted nonparametric hazards approach.

\begin{figure}
\centering
\includegraphics[width=1\linewidth]{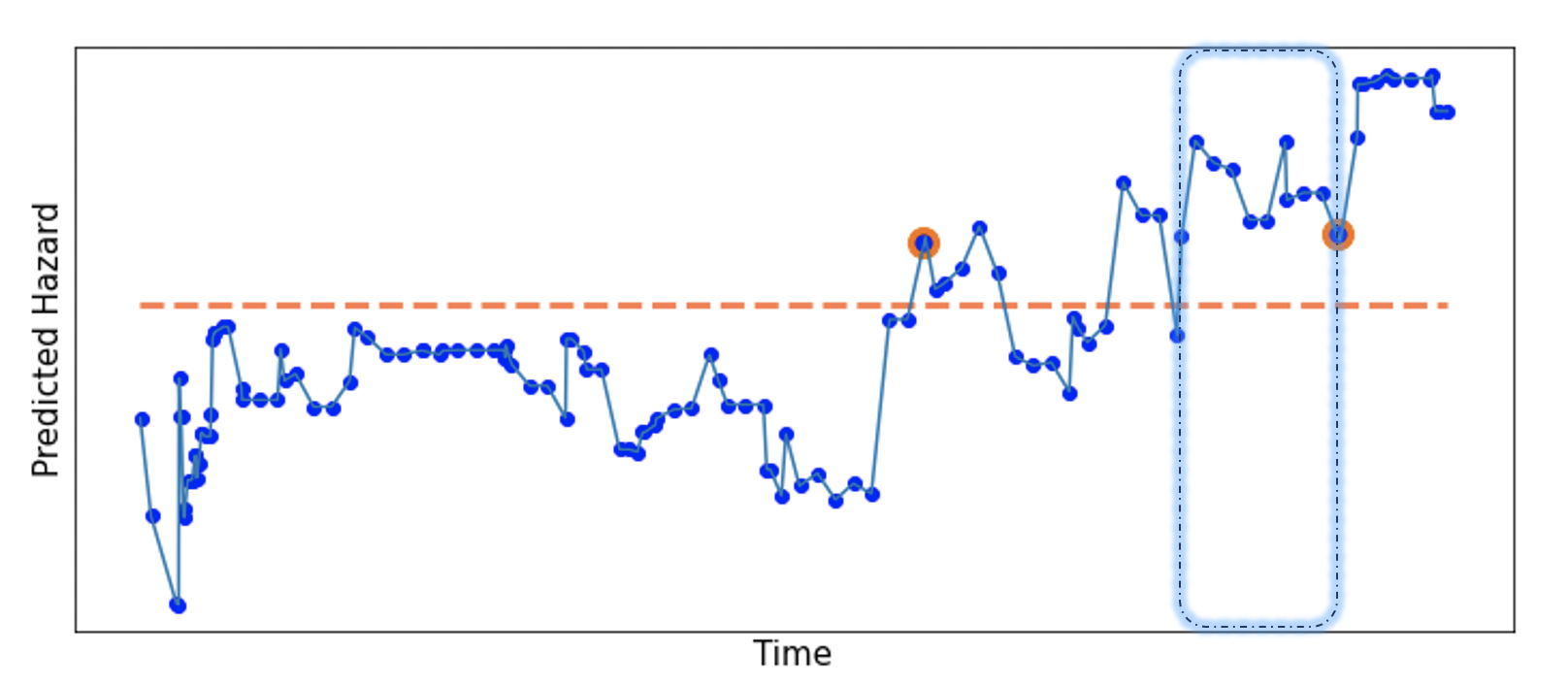}
  \caption{Generating real-time mortality predictions from the values of a patient's risk measure (blue). The first orange dot marks the first time the risk measure exceeds the threshold (dashed horizontal line). The second orange dot marks the first time the risk measure stayed above the threshold for 8 hours (rectangle window).}
  \label{fig:sliding_window}
\end{figure}

For BoXHED, real-time mortality predictions are generated from the values of the risk measure $\hat\lambda(t,X(t))$ over time in the following way (the same approach applies to the risk measures produced by the other two methods): For a given risk threshold $\rho$, we look at whether the patient's risk measure is above $\rho$ for the past 8 hours. A patient is then flagged as predicted to eventually die in the ICU if this is true (as illustrated by the second orange dot in Figure \ref{fig:sliding_window}). Once flagged, no further predictions are made for the patient in question. A second criterion is to flag a patient as soon as the risk measure exceeds $\rho$ (as illustrated by the first orange dot in Figure~\ref{fig:sliding_window}). Further details are provided in the following subsections.

\subsection{Data}

Due to structural differences between the MIMIC III and MIMIC IV datasets, we extract the MIMIC IV dataset using a modified version of the preprocessing pipeline introduced in \cite{harutyunyan2019multitask} for MIMIC III. The modified pipeline merges MIMIC IV’s various tables in order to derive patient ICU history. This results in 31,544 ICU stays, of which two are removed as outliers as they have many more measurements compared to others. All told, this paper focuses on 31,542 ICU stays.

Furthermore, we focus on only the first 120 hours (5 days) of each ICU stay for two reasons. First, the required computational effort for Dynamic DeepHit explodes for a large number of discrete time periods. Second, early intervention is significantly associated with positive patient outcomes \cite{song2012early}, which makes real-time monitoring particularly valuable during the first few days. 
After the initial period, overall patient status is better understood.

The 17 predictive features used to fit our models come from \cite{harutyunyan2019multitask}: Capillary refill rate, Diastolic blood pressure, Fraction inspired oxygen, Glascow coma scale eye opening, Glascow coma scale motor response, Glascow coma scale total, Glascow coma scale verbal response, Glucose, Heart rate, Height, Mean blood pressure, Oxygen saturation, Respiratory rate, Systolic blood pressure, Temperature, Weight and pH. Since these features can be found in many datasets, this makes it easy to compare our results to studies based on other datasets.

\subsection{Training}
The 31,542 ICU stays are randomly split into training and testing sets according to a 80/20 split of the unique patient IDs. This is because one patient may contribute to multiple ICU stay records, so we need to avoid assigning such a patient's second ICU stay to the training set and their first stay to the testing set. All three methods are fit to the training set.

For training BoXHED 2.0, we use the built-in $K$-fold cross-validation function (with $K=5$) to select the number of trees and the depth of the trees to use. Using the one-standard-error rule (\S 7.10 of \cite{hastie2009elements}) to select the most parsimonious model within one standard error of the best performing one, we arrive at using 75 trees of maximum depth 2 (i.e. 4 leaf nodes max).

For training the time-varying Cox model, we use the Lifelines package \cite{cameron_davidson_pilon_2021_4683730} in Python. Unlike BoXHED 2.0 and Dynamic DeepHit, this package does not automatically handle missing data. We therefore impute missing values in the same way as \cite{harutyunyan2019multitask}: If a previous measurement exists, its value is carried forward. Otherwise, the missing feature is imputed using a pre-defined value.

\subsection{Scoring the predictions}
As explained earlier, the thresholding criterion used to flag in-ICU mortality for stays in the testing set is continuously assessed as new data stream in. If the flag is raised during the first 120 hours of the stay, a positive prediction is made for the stay. Otherwise, a negative prediction is made. This approach converts the time-varying output from a dynamic survival model (i.e. the risk measure) into a classification signal. To compute the area under the receiver operating characteristic curve (AUC-ROC) and the area under the precision recall curve (AUC-PRC) for the testing set predictions, the threshold $\rho$ is varied to trace out both curves.

\section{Results}\label{sec:results}
Table~\ref{tab:comparison_single_risk} presents the out-of-sample performances for mortality predictions triggered by the risk measure exceeding the threshold at any point in time. As a reminder, the AUC-ROC baseline of 0.50 corresponds to a random guess, and the AUC-PRC baseline of 0.09 corresponds to always predicting positive. While AUC-ROC is commonly used to evaluate classifiers, AUC-PRC is more informative here (and often in clinical datasets) given the imbalance between the number of negative and positive outcomes \cite{saito2015precision}.\footnote{The benchmark classical prediction model for the decompensation task in MIMIC, the closest to our use case, achieves an AUC-PRC of 0.34 \cite{harutyunyan2019multitask}.} 

We see from Table~\ref{tab:comparison_single_risk} that BoXHED handily outperforms both Time-varying Cox and Dynamic DeepHit, particularly on AUC-PRC. However, a caveat is required for the seemingly dismal performance of Dynamic DeepHit. This could be due to the fact that the method assumes that mortality must occur by some time $\tau_{\max}$ in the ICU, which is inherently incompatible with the current application since 91\% of stays do not end in death. Another possible explanation is suboptimal hyperparameter tuning, which is not as systematic for deep learning as it is for BoXHED. Indeed, tuning neural nets is an art as there are far more degrees of freedom, all the way up to modifying the network architecture itself.

\begin{figure}
\centering
\includegraphics[width=0.65\linewidth]{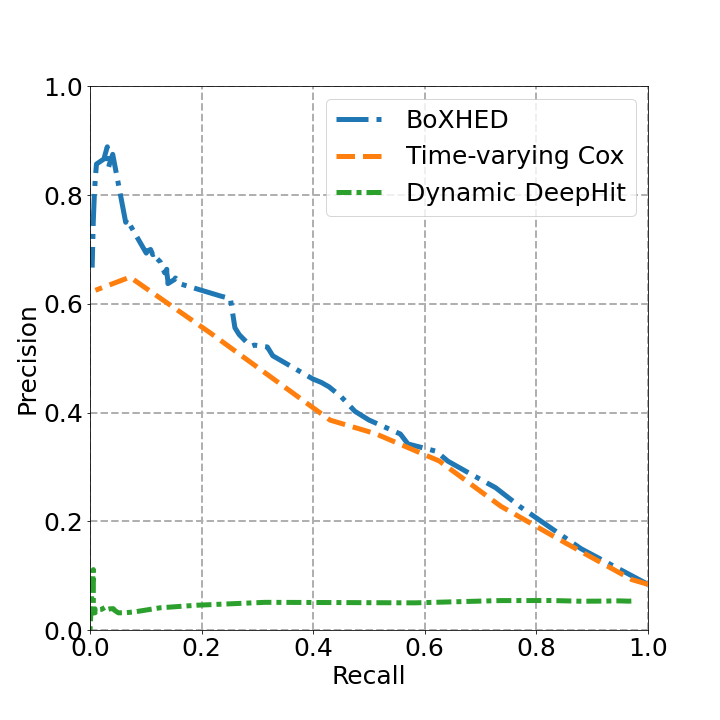}
  \caption{Precision-Recall curve for risk exceeding threshold for entirety of past 8 hours}
  \label{fig:PRC}
\end{figure}

Table~\ref{tab:comparison_sliding_window} presents the out-of-sample performances for predictions based on having the risk measure remain above the threshold for 8 hours. Figure~\ref{fig:PRC} illustrates the precision-recall curve for this case. The direction of the results are qualitatively the same as those in Table~\ref{tab:comparison_single_risk}, except that Time-varying Cox and BoXHED's performances are noticeably better. This is intuitive, since requiring the risk measure to remain elevated for a longer period reduces the number of false positives.

\begin{table}[!t]
\caption{Comparison of model performances: Predictions based on risk measure exceeding threshold at any time (*See \S~\ref{sec:results} for discussion of Dynamic DeepHit results)}
\label{tab:comparison_single_risk}
\begin{center}
\begin{tabular}{ccccc} 
\hline
Model & AUC-ROC & AUC-PRC \\
\hline
Baseline  & 0.50 & 0.09 \\
Time-varying Cox  & 0.74 & 0.29\\
Dynamic DeepHit* & 0.50 & 0.06 \\
BoXHED &\textbf{0.78} & \textbf{0.35} \\
\hline
\end{tabular}
\end{center}
\end{table}

\begin{table}[!t]
\caption{Comparison of model performances: Predictions based on risk measure exceeding threshold for 8 hours (*See \S~\ref{sec:results} for discussion of Dynamic DeepHit results)}
\label{tab:comparison_sliding_window}
\begin{center}
\begin{tabular}{ccccc} 
\hline
Model & Window Size & AUC-ROC & AUC-PRC \\
\hline
Baseline & - & 0.50 & 0.09 \\
Time-varying Cox & 8hrs & 0.81 & 0.36\\
Dynamic DeepHit* & 8hrs & 0.47 & 0.05 \\
BoXHED& 8hrs & \textbf{0.83} & \textbf{0.41}\\
\hline
\end{tabular}
\end{center}
\end{table}

\section{Limitations and Future Work}

Rather than focus on the most recent 8 hours of risk measure values, a moving average might capture more information. Future work could explore the potential of flagging patients if the moving average exceeds some threshold. Second, the current prediction target is whether or not a patient will eventually die in the ICU, be it in 2 hours or 2 days. To help physicians prioritize care for patients at higher imminent risk, future research should consider shorter prediction horizons such as in-ICU  death within 6 hours \cite{ma2019shapes}. Lastly, the outcome of interest should be expanded from in-ICU mortality to in-hospital mortality, and to account for competing risks from multiple adverse events, potentially over different time scales.

\section{Conclusion}
EHR data is a rich source of information for adverse event prediction in clinical settings. The high-frequency, time-varying data present opportunity to develop real-time warning systems that update estimates of patient mortality hazards with the introduction of each new data point. Survival analysis is the ideal tool for this. However, there is a dearth of survival methods that can handle time-varying features nonparametrically and at scale. This work presents the application of such a tool called BoXHED for developing an in-ICU mortality warning system using MIMIC-IV data. The system achieves state-of-the-art results (AUC-ROC 0.83, AUC-PRC 0.41) when compared to the benchmarks. The results highlight the promise of BoXHED, a gradient-boosted nonparametric hazard estimator, for real-time clinical predictions.

\bibliographystyle{IEEEtran}
\bibliography{IEEEabrv,refs_boxhed}

\end{document}